\documentclass{article}

\usepackage[preprint, nonatbib]{tsrml_2022}

\usepackage[utf8]{inputenc} 
\usepackage[T1]{fontenc}    
\usepackage{hyperref}       
\usepackage{url}            
\usepackage{booktabs}       
\usepackage{amsfonts}       
\usepackage{nicefrac}       
\usepackage{microtype}      
\usepackage{xcolor}         
\usepackage{graphicx}
\usepackage{wrapfig}

\usepackage[numbers]{natbib}
\setcitestyle{numeric}

\usepackage[utf8]{inputenc} 
\usepackage[T1]{fontenc}    
\usepackage{hyperref}       
\usepackage{url}            
\usepackage{booktabs}       
\usepackage{amsfonts}       
\usepackage{nicefrac}       
\usepackage{microtype}      
\usepackage{xcolor}         

\usepackage{wrapfig}
\usepackage{multirow}
\usepackage{tabularx}
\usepackage{caption}
\usepackage{graphicx}

\title{Interpreting Bias in the Neural Networks: A Peek Into Representational Similarity}

\author{%
   Gnyanesh Bangaru* \\
   VNRVJIET \\
   Hyderabad-90, T.S, India \\
   \texttt{cgnyanesh@gmail.com} \\
      \And
  Lalith Bharadwaj Baru\thanks{These authors have equally contributed and \emph{Lalith} has done this work while studying at VNRVJIET.} \\
  IHub-Data, IIIT Hyderabad\\
  Hyderabad-08, T.S, India \\
  \texttt{lalithbharadwaj313@gmail.com} \\
   \And
   Kiran Chakravarthula* \\
   VNRVJIET \\
   Hyderabad-90, T.S, India \\
   \texttt{kiran\_c@vnrvjiet.in} \\
}

\begin{document}

\maketitle

\begin{abstract}
Neural networks trained on standard image classification data sets are shown to be less resistant to data set bias. It is necessary to comprehend the behavior objective function that might correspond to superior performance for data with biases. However, there is little research on the selection of the objective function and its representational structure when trained on data set with biases.

In this paper, we investigate the performance and internal representational structure of convolution-based neural networks (e.g., ResNets) trained on biased data using various objective functions. We specifically study similarities in representations, using Centered Kernel Alignment (CKA), for different objective functions (probabilistic and margin-based) and offer a comprehensive analysis of the chosen ones. 

According to our findings, ResNets representations obtained with Negative Log Likelihood $(\mathcal{L}_{NLL})$ and Softmax Cross-Entropy ($\mathcal{L}_{SCE}$) as loss functions are equally capable of producing better performance and fine representations on biased data. We note that without progressive representational similarities among the layers of a neural network, the performance is less likely to be robust.
\end{abstract}

\section{Introduction}

Deep neural networks that have been trained with a huge amount of data pertaining to multiple classes have provided the most robust performance. Techniques like Domain-Adaptation and Transfer Learning are meant to be widely used, as they help to derive better transferable features to gain considerable performance. The prime objective of a neural network is to interpret the data over providing accurate performance and whereas training these neural networks in a controlled setting \cite{russin2020deep} leads them to perform poorly on unseen data, especially in the case of data with biases.

Moreover, the usage of pre-trained models produces biased results as the sought-after ImageNet tends to learn certain gender, racial and multiple inter-sectional biases as addressed by Kate Crawford \textit{et al.} \cite{Crawford2021} and Ryan \textit{et al.}  \cite{Steed2021ImageRL}. A study, addressed in this work \cite{eitel2021beyond}, states that by the present time, 85\% of AI projects tend to produce erroneous results due to the presence of bias and loss of fairness in the existing data and algorithms. The learning progression of deep learning paradigm is turning to baffle model’s interpretability and make the models amiss.

Fortunately, certain optimization techniques for neural networks are been designed to address the representational \emph{bias} problem. Further, an evaluation strategy for comprehending the relationship between feature spaces of multiple dimensions has been developed \cite{asi2021stochastic},  \cite{li2019repair}, \cite{Zhen2022OnTV}.

The representations learned through these neural networks are predominant towards providing prodigious performance whereas they are concealed w.r.t to their internal layer-based interpretation. This majorly leads the models to under perform on such data with biases. 

Specifically, It is thus very crucial to interpret the reason behind such a learning mechanism to tackle the problem of bias. In this process, we address the following questions: What may be an ideal objective function for the biased data? What sort of representations do neural networks learn in due course, after being exposed to such data? How are the internal representations of neural networks impacted by altering the data samples (\emph{align} and \emph{conflict})? Finally, how intrepretable are the representations that are trained using biased data?

\paragraph{Contributions} The contributions of this work are listed as,
\begin{enumerate}
    \item Our Analysis suggests that $\mathcal{L}_{NLL}$, and $\mathcal{L}_{SCE}$ objective functions could contribute for a best performance with data with biases.
    \item We have illustrated the internal representational structure of each model using CKA and conclude that, training a neural network with $\mathcal{L}_{NLL}$ or $\mathcal{L}_{SCE}$ loss attained better performance with progressive similarity across the layers. 
\end{enumerate}

\begin{wraptable}{r}{9cm}
  \centering
  \begin{tabular}{@{}|c|c|c|c|@{}}
    \hline
    Dataset Kind& Dataset Name& Classes & Train-Val-Test  \\
    \hline
    Biased & C-MNIST & 10 & 55k-5k-10k \\
    Biased & CIFAR-C & 10 & 45k-4.8k-10k \\
    Biased & B-FFHQ & 2 & 19.2k-1k-1k \\
    \hline
  \end{tabular}
  \vspace{4pt}
  \caption{The above table illustrates data sets containing biased samples. 'NA' is specified to denote the absence of samples for that particular split.}\label{tab:1}
\end{wraptable}

\section{Related Works} 
Simon \textit{et al.} \cite{kornblith2019similarity} provides deeper insights into the transferability of representations when exposed to various loss functions, but this work majorly focuses on probabilistic objectives. Kim\textit{ et al.} \cite{kim2019learning} proposes a novel regularized loss function to unlearn the target bias in the data, based on mutual information obtained from InfoGANs \cite{chen2016infogan}, and a gradient reversal layer. Over time, it helps in reducing the pernicious effects of bias in data. Adeli \textit{et al.} \cite{adeli2021representation} proposes a dual-objective adversarial training loss function to learn traits that show the least statistical dependence and the maximum percipience with protected bias. We first examine the empirical performance of the objective functions by classifying them into two groups: (a) probabilistic, and (b) margin-based variations. Later, we examine the significance of particular objective functions that offer representation structure with strong generalizability and transferability.

\section{Setup}
\paragraph{Terminology}
The input data is represented as $X \in \{ x^{(1)},x^{(2)}, ..., x^{(n)}  \}$ where $n$ is the total number of samples. The encoder $Enc(.)$ is used to extract features from a given input $X$. The features extracted from the encoder are presented as $fv \gets Enc(X)$; where $fv \in \{ fv^{(1)}, fv^{(2)}, ..., fv^{(n)}\}$. The dimensions of the feature vector vary by changing the encoder. As most of the experiments were carried out using a supervised framework, the data sets do have certain ground truth labels, and these are represented as $Y \in \{ y^{(1)},y^{(2)}, ..., y^{(n)} \} $. The activation functions, sigmoid and softmax, are indicated by $\sigma(a_i) = 1/ (1+e^{-a_i})$ and $\mathcal{S}(a_i) = e^{a_i}/\sum_j e^{a_j}$, respectively. The loss (objective) function, is denoted by $\mathcal{L}_{(.)}$ and the suffixes indicate its specified variant. The norms $\| . \|_1$ and $\| . \|_2$  indicate Manhattan and Euclidean norms\footnote{Suppose, $a \in \mathbb{R}^n$ then, $\|a\|_1 = \sum_i^{n} |a_i|$ ; $\|a\|_2 = \sqrt{\sum_i^{n} a_i^2}$} respectively.

\paragraph{Models}
In this work, we use ResNet18 to understand convolution-type representations. ResNet18 is used for all of the data, which are illustrated in Table~\ref{tab:1}. We have considered standard ResNets with global average pooling. The fully connected layers for ResNet18 are [512-$c$] (Where $c$ denotes the number of classes). Intermediate dropout layers are used with a drop rate of 40\% and a set constant for all data sets to have a fair evaluation.

\paragraph{Training and fine-tuning}
To train each model we utilized Adam \cite{kingma2014adam} as an optimizer with a standard learning rate of $10^{-3}$ and a weight decay of $10^{-5}$. For all of the executions, we trained each model from scratch. We fed 512 samples in batches to neural networks by varying the objective functions. The early stopping criterion is embedded with the patience of 12 epochs to ensure the model does not over-fit with excessive training. The results reported in Tables \ref{tab:2} are produced without any augmentations (except normalisation) and pre-trained weights.

\paragraph{Datasets}
As mentioned, for bias data classification, we have utilized the data set provided by Lee \textit{et al.} \cite{lee2021learning} where they aim to solve the bias problem by providing 3 different biased datasets: Colored MNIST, Corrupted CIFAR, and Biased FFHQ. From the provided datasets, we have analysed all of these datasets. Each data set consists of various diversity ratios in order to tackle the problem of bias. In which, bias is reduced by providing diverse bias-conflicting samples i.e, \textit{align} and \textit{conflict} divisions in the data set. The partition of data \textit{conflict} is based on a percentage of diversity and we have used 5\% diversity, among the varying diversity ratios. The number of diverse samples in that particular bias-conflicting data set is more compared to that of 0.5\% and 1\% i.e., in colored MNIST bias-conflicting samples are considered to have more images with differently colored digits whereas in 0.5\% and 1\%  we have less diversity of bias-conflicting samples. 

\begin{wraptable}{r}{7cm}
\begin{center}
\scalebox{0.60}
{
\begin{tabular}{|c|c|}
\hline
Loss & Equation \\
\hline\hline
$\mathcal{L}_{SCE}$ & $(fv^{(i)}, y^{(i)})  = - \sum_{c=1}^C y_c^{(i)}\texttt{ log}\left(\mathcal{S}(fv_c^{(i)})\right)$ \\

$\mathcal{L}_{BCE}$&$ (fv^{(i)}, y^{(i)})   = y^{(i)} \texttt{log}(\sigma(fv^{(i)}))  + (1- y^{(i)}) \texttt{log}(1 - \sigma(fv^{(i)}))$ \\
$\mathcal{L}_{NLL}$& 
$(fv^{(i)}, y^{(i)})   = - \sum_{c=1}^C y_c^{(i)}\texttt{ log}\left(fv_c^{(i)}\right)$\\
\hline
$\mathcal{L}_1$& 
$ (fv^{(i)}, y^{(i)})  =  - \sum_{c=1}^C \| y_c^{(i)} - \mathcal{S}(fv_c^{(i)}) \|_1$ \\
$\mathcal{L}_2$& $(fv^{(i)}, y^{(i)})  =  - \sum_{c=1}^C \| y_c^{(i)} - \mathcal{S}(fv_c^{(i)}) \|_2 ^2 $\\
$\mathcal{L}_{SoS}$& $(fv^{(i)}, y^{(i)}) =      \frac{1}{C}{ \sum_{c=1}^C \left[\alpha y_c^{(i)} (fv_c^{(i)}-\beta) + (1 - y_c^{(i)}) (fv_c^{(i)})^2 \right] } $\\
\hline
\end{tabular}
}
\end{center}
\caption{The above table illustrates all the objective functions that are experimented with in this work.}\label{tab:loss}
\end{wraptable}

\section{Experimentation}
The choice of the objective function to train a deep neural network, on specified data remains a question. Janocha \textit{et al.} \cite{janocha2017loss} provided a theoretical justification and conducted experiments on MNIST pointing out the importance of $\mathcal{L}_1$ and $\mathcal{L}_2$ not just as regularizers, but as objective functions for better generalisations. Hui \textit{et al.} \cite{hui2021evaluation} empirically proves that square loss with a little parametric tuning would produce significant results for most tasks of natural language processing (NLP) and automatic speech recognition (ASR). Hui \textit{et al.} \cite{hui2021evaluation} specifically mentioned that the proposed square loss is not brittle for randomised initialisation. A recent analysis by Simon \textit{et al.} \cite{kornblith2021better} provides insights noting that the representations acquired to classify certain tasks with more class separation lead to poor transferable features. This work implies various objective functions to observe both the performance and quality of representations for the standard computer vision classification data sets. 

\begin{figure*}[t!]
\centering
\includegraphics[width=0.99\textwidth]{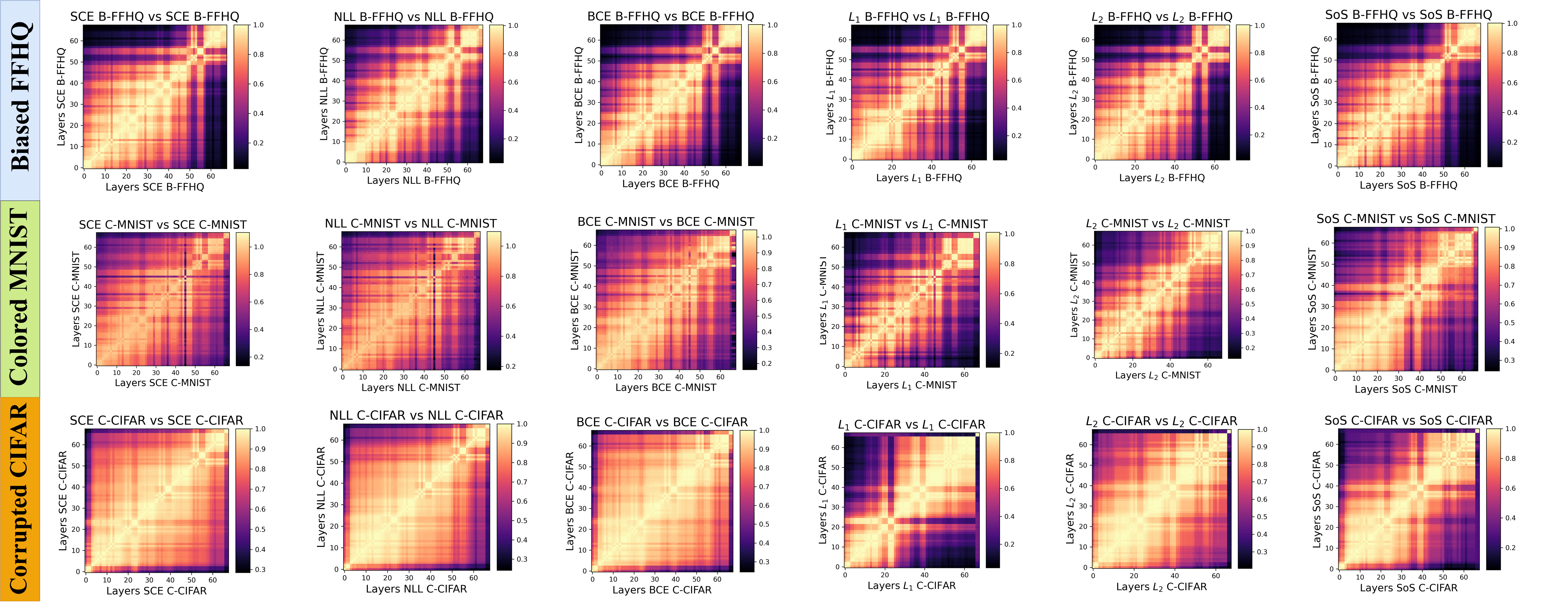}
\caption{This figure illustrates CKA visualizations for all the objective functions on B-FFHQ, C-MNIST, and C-CIFAR data.  Each tile of the plot consists of a color, indicating the correlation of representations i.e., the similarity of each layer representation. For these, corresponding color map is provided on the right side of each image tile. For visualizing CKA, we considered all the layers of the neural network (ResNet18) including activation, normalization, and fully-connected layers. The matrix is formed by comparing the features acquired by each layer of the ResNet18 with itself.}\label{fig:x}
\end{figure*}

The previous literature focuses on the training and transferability of features acquired by training standard neural architectures with varying objective functions. But, there is sparse literature noting the relevance of both probabilistic and margin-based objective functions on data with biases and distributional shifts. Hence, we provide empirical analysis for two variants of objective functions to understand the performance of each objective function on various data sets mentioned in Table \ref{tab:1}.

\subsection{Probabilistic Objectives}
Probabilistic objective functions calculate the error that approximates the underlying probabilities for representations acquired from an encoder (ResNet). In this paper, we include three probabilistic objectives and they are detailed in Table \ref{tab:loss}. First, the Softmax cross-entropy~\cite{bridle1989training} ($\mathcal{L}_{SCE}$), a highly used objective,  is obtained by applying softmax activation in the final layer of the neural network, and this feed is minimised by the negative log-likelihood ($\mathcal{L}_{NLL}$). Next, Binary cross-entropy ($\mathcal{L}_{BCE}$) is obtained by applying sigmoid activation ($\sigma(.)$) at the final layer of neural networks and this information is minimised by NLL. 

Primarily, the $\mathcal{L}_{BCE}$ loss function is used for binary classification problems but, a recent work empirically proves that its implication on multi-class would lead to better performance \cite{beyer2020we} by applying the one-vs-rest strategy. Finally, the likelihood provides the joint probability of the sample distribution and minimises the negative logarithm of the obtained likelihood \cite{bishop1995neural}.

\subsection{Margin-based Objectives} 
Margin-based objective functions calculate the error by discriminating the representations extracted from an encoder (ResNet). Similarly to probabilistic objectives, we include three margin-based objectives, which are detailed in Table \ref{tab:loss}. First, the mean absolute error ($\mathcal{L}_1$ objective function) finds the Manhattan distance between the two representations. We acquire the theoretical motivation of Janocha \textit{et al.} \cite{janocha2017loss} that $\mathcal{L}_1$ would reduce the sparseness in the representations. Rather than directly discriminating the representations in the final layer, we use softmax to ensure appropriate learning without saturation of partial derivatives.

Similar to $\mathcal{L}_1$, we use $\mathcal{L}_2$ to find find the Euclidean distance between two representations (final layer). Lastly, Hui \textit{et al.} \cite{hui2021evaluation} rescaled SoS
to be more robust by providing two parameters $\alpha$, $\beta$. Injection of these parameters resulted in a decent performance for the
NLP and ASR tasks, but was poorly performed on the computer vision tasks. For experimentation, we have chosen $\alpha$, $\beta = 1$, 
and this reduces to standard SoS.

\subsection{Empirical Analysis}
Now, let us understand the empirical performance of these objective functions trained on all variants of the data detailed in Table \ref{tab:2}. A detailed mode of training, choice of neural networks, data sets and hyperparameters are neatly detailed in the Setup section.In most of the cases, $\mathcal{L}_{NLL}$ was able to achieve top accuracy scores. But, $\mathcal{L}_{SCE}$ obtained the highest accuracy for B-FFHQ and $\mathcal{L}_{2}$ competed closely with it. Taking into account the case of all of the biased data  $\mathcal{L}_{SCE}$ performed standalone; $\mathcal{L}_{NLL}$ was able to compete closely with $\mathcal{L}_{NLL}$. Hence, aggregating these results, it is strongly recommended that using probabilistic loss functions ($\mathcal{L}_{SCE}$ and $\mathcal{L}_{NLL}$) to obtain decent performance on most of the biased data. The empirical performance attained by these objective functions is well-understood but, the question arises with the internal representations of the model trained on these data and for this we comprehend the underlying the representational structure.

\subsection{Representational Analysis}
 The \emph{Centered Kernel Alignment (CKA)} was devised to understand the representation structure of artificial neural networks. It is well established in the literature that CKA \cite{kornblith2019similarity} acquires qualitative representations compared to PwCCA \cite{morcos2018insights} and SVCCA \cite{raghu2017svcca}. CKA not only captures the correspondence between the representations of a neural network but also allows us to compute the similarity between pairs of layers. 
 

To reduce the computational expense consumed by linear CKA, mini-batch CKA is applied by computing the mean of HSIC (Hilbert-Schmidt Independence Criterion) scores on selected mini-batches ($N$). This strategy is implemented straightforwardly as Thao et al. \cite{nguyen2021wide}. The mini-batch CKA is detailed as follows: 

\begin{equation}
    \texttt{CKA}^{mini} = \frac{\sum_{i=1}^N \texttt{HSIC}_i(\tilde{X_i}, \tilde{Y_i})}{\sqrt{\sum_{i=1}^N \texttt{HSIC}_i(\tilde{X_i}, \tilde{X_i})} \sqrt{\sum_{i=1}^N \texttt{HSIC}_i(\tilde{Y_i}, \tilde{Y_i})} }
\end{equation}

where, $\tilde{X_i} = X_i X_i^T; \tilde{Y_i} = Y_i Y_i^T$. These $X_i \in \mathbb{R}^{n \times d_1}$ and $Y_i \in \mathbb{R}^{n \times d_2}$ are activation matrices for i$^{th}$ mini-batch of examples without replacement. We now try to analyse the representations using CKA for all the objective functions on the aforementioned data\footnote{To understand the CKA and its underlying significance we request the readers to go through the works \cite{kornblith2019similarity, nguyen2021wide}.}.

Now we aim to address, which layers correspond to similar representations in a specific neural network trained on a certain objective function. In the following, we are going to understand which representations would lead to better outcomes and which do not. For this, we intend to choose all the objective functions for representation analysis using CKA and the biased data variants mentioned in Table \ref{tab:1}. 

In Figure \ref{fig:x}, when considering B-FFHQ data, the CKA representation matrix formed for $\mathcal{L}_{NLL}$, $\mathcal{L}_{BCE}$ and $\mathcal{L}_{SCE}$ seems to have similar characteristics. While considering the case for the C-MNIST data set, all the loss functions tend to form a small box-like structure at the ultimate layers (at the top right corner after 50$^{th}$ layer). But, $\mathcal{L}_{NLL}, \mathcal{L}_{SoS},$ and $\mathcal{L}_{SCE}$ seem to have uniformly distributed representation with decreasing similarity with the depth of the neural network. Finally, for C-CIFAR data the refined representation similarity is obtained for $\mathcal{L}_{SCE}, \mathcal{L}_{BCE}$ and  $\mathcal{L}_{2}$. 

However, the performance for all the data is higher for Probabilistic objectives. Considering the case of $\mathcal{L}_{SoS}$ objective, it is prone to have \emph{block structured} representations on utmost all the data~\cite{nguyen2021wide}. This structured block resembles the neural network as \emph{overparameterised} model. The underlying reason is either that the model has fewer data samples or a deeper network. This can be surmounted by truncating the layers with identical representational similarity.

These indications are clear to note that, the objectives $\mathcal{L}_{SCE}$ and $\mathcal{L}_{NLL}$ not only provide decent empirical performance but capture fine representations with ResNets. Hence, from this interpretation we infer that, representations acquired by probabilistic objectives are comparatively better to provide good generalisations for diverse bias data sets. 


\begin{table*}[t!]
\begin{center}
\scalebox{1.05}{
\begin{tabular}{ |c|c|ccc|c| } 
\hline
\multirow{2}{6em}{\textbf{Objectives}} & \multirow{2}{4em}{\textbf{Variants}} & \multicolumn{3}{c|} {\textbf{Biased}} & 
\multirow{2}{4em}{ \textbf{Mean} }\\ 
  & & C-MNIST & C-CIFAR & B-FFHQ& \\ \hline\hline
  \multirow{3}{6em}{\textbf{Probabilistic}}  
  &$\mathcal{L}_{SCE}$	&	\underline{95.31$\pm$ 1.21} &	34.44$\pm$ 1.84 &\bf 57.03$\pm$2.42
  & {\bf {62.26}}\\
  
  &$\mathcal{L}_{BCE}$	&	93.75$\pm$ 1.67 &	32.47$\pm$ 4.34 & 54.40$\pm$1.80
  &60.20\\
  
  &$\mathcal{L}_{NLL}$	&\bf	95.81$\pm$ 1.36 &	\bf 35.25$\pm$ 1.79 & 53.19$\pm$1.79
  &\underline{61.42}\\
  \hline

\multirow{3}{7em}{\textbf{Margin-based}} 
&$\mathcal{L}_1$	&	95.19$\pm$ 2.64 &25.11$\pm$ 1.21 & 52.53$\pm$1.80
&57.61\\

&$\mathcal{L}_2$	&	93.00$\pm$ 0.99 &	 25.34$\pm$ 2.64 & \underline{55.73$\pm$1.57}
&58.02 \\

&$\mathcal{L}_{SoS}$	&	95.00$\pm$ 0.63 &	\underline{34.81$\pm$ 1.42} & 49.69$\pm$2.54
&59.83 \\ 
\hline
\end{tabular}
}
\vspace{6pt}
\caption{The table below provides the empirical performance of the individual objective functions for the generic, bias, and OOD data. The experiments were carried out without any augmentations and did not use learnt weights (ImageNet1K or ImageNet21K) for the training models. These experiments were conducted three times for each objective function for a fair evaluation. The tabulated mean and standard deviation (mean $\pm$ std) in each cell depicts the accuracy scores obtained after experimenting thrice with the 'test' data. \textbf{Bold} and \underline{underline} represent the accuracy scores of \textbf{first} and \underline{second} best performing models, respectively.}\label{tab:2}
\end{center}
\end{table*}

\vspace{20pt}

\section{Conclusion}
By summarising the above, we infer that $\mathcal{L}_{NLL}$ and $\mathcal{L}_{SCE}$ objectives would be apt for data with biases. But while experimenting, it should be noted that the variance in accuracy must be minimalist to ensure robustness. Next, if the neural networks are exposed to biased data, generalisation is attained only when the layers of CKA matrices have a progressive dissimilarity with the depth of the network. 

In the future, we see the potential requirement of data sets comprising samples with biases as big as ImageNet. Similarly, the representations acquired by the models are to be ensured with the least bias possible. We believe that, comprehending the representations acquired from biased data would aid researchers in providing a novel debiasing neural network or a bias mitigation strategy. Also we believe that, this CKA framework can be extended to study the detailed intrepretability of the neural networks \cite{kar2022interpretability}

\section{Ethical Aspects and Broader Impact}
Inherent biases of the individuals who train AI models inevitably reflect in the models. Such inherent biases are due to psychological constructs, different types of biases, and the inter-working thereof \cite{eitel2021beyond}. In addition to the unintentional, implicit, and indirect nature of these inherent biases, the need to retain certain biases—such as demographic aspects of the data—needed to qualify the data increases the complexity in the overall bias. Unconventional training and evaluation methods may be necessary to de-bias the AI models and minimize human interventions, leading to the design and development of meta approaches to create ethical or responsible AI systems.

Since scientists and analysts routinely conclude, albeit based on evidence-based insights, biased data can be inherently misleading. The impact of such misinterpretation can be corrected through exposure, experience, and expertise when humans are involved in the interpretation. However, if the data itself is biased or only the interpretations of the data were presented by researchers, the danger of unconscious or even conscious biases cannot be ruled out entirely. While Smith's book \cite{smith2021decolonizing} is an elaborate example of how research methodologies are inherently designed with ignorance towards the subjects of study, subtler cases of bias can be innate to the ways data are sourced, stored, pre-processed, analyzed, and interpreted. 

The current work emphasises the need to methodically improve the quality biased representations obtained from neural networks without proposing radical paradigm shifts in current methodologies. This work does not organically provide scope for misinterpretation of data or biased decisions made through data analytics. Furthermore, the work facilitates a better and more uniform representation of the data by reminding researchers to consciously consider the biased aspects of the data, which may be rather inconspicuous. A stronger motivation arises as Artificial Intelligence and Machine Learning continue to be used in various technology and social domains for diverse applications.

{
\small
\bibliographystyle{plain}
\bibliography{output}
}




\end{document}